\documentclass[times, review, 10pt]{elsarticle}

\usepackage{amssymb}
\usepackage{amsmath}
\usepackage{graphicx}
\usepackage{subfigure}
\usepackage{multirow}
\usepackage{booktabs}
\usepackage{hyperref}
\usepackage{color, xcolor}
\usepackage{algorithm}
\usepackage{algorithmicx}
\usepackage{makecell}
\usepackage{verbatim}

\journal{Information Fusion}

\begin{document}

\begin{frontmatter}



\title{RVAFM: Re-parameterizing Vertical Attention Fusion Module for Handwritten Paragraph Text Recognition} 


\author[address1]{Jinhui Zheng}
\ead{zhengjinhui95@stu.jnu.edu.cn}

\author[address2]{Zhiquan Liu}
\ead{zqliu@vip.qq.com}
\author[address3]{Yain-Whar Si}
\ead{fstasp@um.edu.mo}
\author[address4]{Jianqing Li}
\ead{jqli@must.edu.mo}
\author[address1]{Xinyuan Zhang}
\ead{zhangxy@jnu.edu.cn}
\author[address1]{Xiaofan Li}
\ead{lixiaofan@jnu.edu.cn}
\author[address1]{Haozhi Huang}
\ead{hzhuang@jnu.edu.cn}
\author[address1]{Xueyuan Gong\corref{cor1}}
\ead{xygong@jnu.edu.cn}
\cortext[cor1]{Corresponding authors.}
\affiliation[address1]{organization={School of Intelligent Systems Science and Engineering},
             addressline={Jinan University},
             state={Guangdong Province},
             country={China}}

\affiliation[address2]{organization={College of Cyber Security},
             addressline={Jinan University},
             state={Guangdong Province},
             country={China}}
             
\affiliation[address3]{organization={Department of Computer and Information Science},
             addressline={Faculty of Science and Technology, University of Macau},
             state={Macau},
             country={China}}
\affiliation[address4]{organization={School of Computer Science and Engineering},
             addressline={Macau University of Science and Technology},
             state={Macau},
             country={China}} 

\begin{abstract}
Handwritten Paragraph Text Recognition (HPTR) is a challenging task in Computer Vision, requiring the transformation of a paragraph text image, rich in handwritten text, into text encoding sequences. One of the most advanced models for this task is Vertical Attention Network (VAN), which utilizes a Vertical Attention Module (VAM) to implicitly segment paragraph text images into text lines, thereby reducing the difficulty of the recognition task. However, from a network structure perspective, VAM is a single-branch module, which is less effective in learning compared to multi-branch modules. 
In this paper, we propose a new module, named Re-parameterizing Vertical Attention Fusion Module (RVAFM), which incorporates structural re-parameterization techniques. RVAFM decouples the structure of the module during training and inference stages. During training, it uses a multi-branch structure for more effective learning, and during inference, it uses a single-branch structure for faster processing. The features learned by the multi-branch structure are fused into the single-branch structure through a special fusion method named Re-parameterization Fusion (RF) without any loss of information. 
As a result, we achieve a Character Error Rate (CER) of 4.44\% and a Word Error Rate (WER) of 14.37\% on the IAM paragraph-level test set. Additionally, the inference speed is slightly faster than VAN.
\end{abstract}




\end{frontmatter}



\section{Introduction}
\label{sec:intro}
Handwritten Paragraph Text Recognition (HPTR) is currently a prominent yet challenging task, which translates images containing numerous handwritten characters in various styles, e.g., letters and draft document images, into text encoding sequences. With the rapid advancement of the information society, the deployment of HPTR models enables the digitization of document images, thereby facilitating the storage and processing of such data using computers. Furthermore, this capability can serve as an upstream process for various other tasks, including document summarization and document analysis.

The Vertical Attention Network (VAN) \citep{coquenet2022end} is presently among the most advanced models for HPTR, which employs Vertical Attention Modules (VAM) to implicitly segment the paragraph into text lines. VAM is a module that employs a vertical attention mechanism. It takes a feature map of a handwritten paragraph text image as input, and the features of each text line as the output in order to achieve implicit segmentation. During this process, VAM calculates the weight for each row in the feature map using both global and local information. It then performs a weighted sum to output the line features for the current stage. This process is carried out from the top to the bottom of the paragraph until the last line is reached. VAM simplifies the task from paragraph text recognition to text line recognition, thereby reducing the overall difficulty of the recognition task. However, because VAM is a single-branch structure, unlike the residual structure in ResNet \citep{he2016deep} and branch-concatenation in Inception \citep{szegedy2015going, szegedy2017inception, ioffe2015batch, szegedy2016rethinking} both that have multi-branch designs, the module's expressive capability is limited. If VAM is only simply modified to a multi-branch structure, it would also lead to a significant reduction in the module's inference speed. One of the main reasons for this dilemma is that the structure during the training stage is often the same as the structure during the inference stage. In this paper, we propose a new vertical attention module called Re-parameterizing Vertical Attention Fusion Module (RVAFM), inspired by Structural Re-parameterization \citep{ding2021diverse}, which decouples the training and inference stages of the module, to solve the aforementioned problem.

\begin{figure*}
    \centering
    \includegraphics[width=1\linewidth]{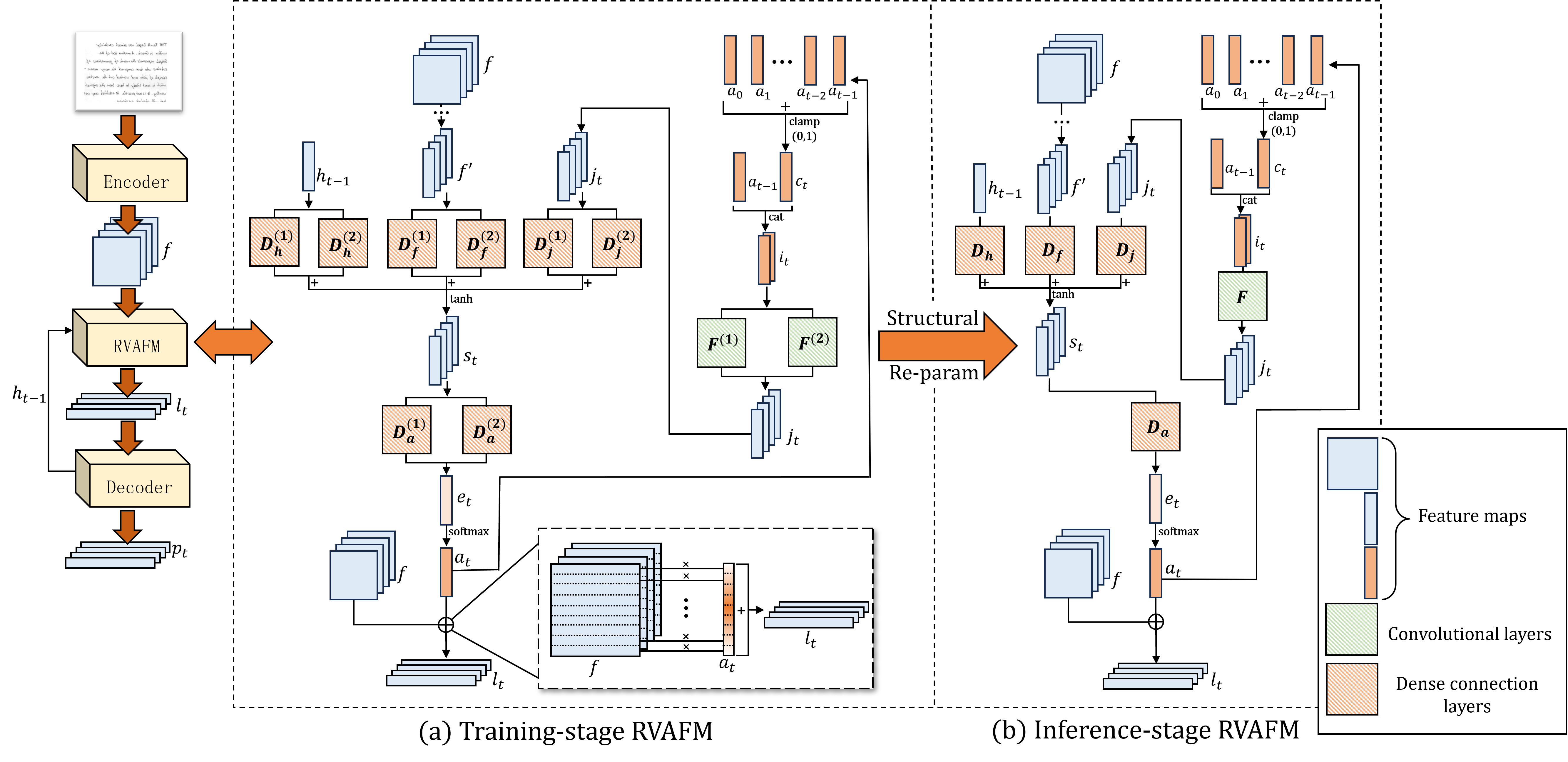}
    \caption{The structure of RVAFM. (During the training process, the Training Stage RVAFM structure is used, which employs dual-parameter layers. After training is completed, we fuse the dual-parameter layers into single-parameter layers through structural re-parameterization, resulting in the Inference Stage RVAFM structure, which is used for inference.) }
    \label{fig:arch-repvamuyb}
\end{figure*}

The overall structure of RVAFM is illustrated in Fig. \ref{fig:arch-repvamuyb}. We employ \textbf{dual-parameter layers} (dual-convolution layers or dual-dense layers) along with a specific fusion method called Re-parameterization Fusion (RF), enabling RVAFM to different structures during the training and inference stage, respectively. During the training stage, RVAFM is a multi-branch structure, with dual-parameter layers. In each dual-parameter layer, the inputs of the two parallel layers are identical, the outputs of them are summed before the next part, which means there are two branches in a dual-parameter layer. RVAFM applies a total of five dual-parameter layers, resulting in a final structure with $2^{5}$ branches. During the inference stage, the features learned by each dual-parameter layer are fused into the corresponding single-parameter layer through RF, which is formulated based on the distributive property of addition for convolution operations and matrix multiplication. Importantly, this fusion process ensures that the accuracy is preserved without any loss. Finally, by the dual-parameter layer, we have differentiated the structure of RVAFM between its training and inference stages. During the training stage, RVAFM features a complex architecture to enhance its ability to express intermediate feature representations and improve learning competence, thereby increasing the overall model's learning capability. During inference stage, RVAFM adopts a simplified structure to optimize the model's inference speed. Simultaneously, through RF, we establish a seamless bridge for module parameters between the training and inference stage. Following experimental validation, 
the VAN incorporating RVAFM (referred to as RVAN) achieves better performance than VAN, while also having a slightly faster inference speed.

In summary, this paper has the following contributions:
    \begin{itemize}
        \item This paper proposes dual-parameter layers to modify VAM, resulting in a new module named RVAFM, which has a multi-branch structure during the training stage. This enables RVAN to adopt a more complex architecture, thereby enhancing its learning capability.
        \item This paper selects a fusion method named RF to fusion of RVAFM's multi-branch structure into a single-branch structure, applying the obtained simplified structure during the inference stage to accelerate speed.
        \item A series of experiments are conducted in this paper, demonstrating that RVAN reaches the state-of-the-art in the field of HPTR. These experiments also prove that RVAN, through its structural re-parameterization design, overcomes the shortcomings of the original model and achieves better performance. Additionally, we investigate the characteristics of RVAN through hyper-parameter tuning experiments and identify the optimal hyper-parameter configuration for RVAN.
    \end{itemize}

The remaining parts of this paper are organized as follows: Related works are discussed in Section \ref{sc:rw}. The details of the proposed architecture are presented in Section \ref{sc:architecture}. Section \ref{sc:experiments} focuses on the experimental environment and a series of experiments related to RVAN. Section \ref{sc:conclusion} wraps up the paper with conclusions.

\section{Related work}
\label{sc:rw}

RVAFM represents an attempt to apply Structural Re-parameterization to the field of HPTR. To clarify the present work, this section will review prior research that is closely related to the fields of HPTR and Structural Re-parameterization.

\subsection{Handwritten Paragraph Text Recognition}
\label{ssc:hptr}
The task of HPTR can also be considered as multi-line text recognition, which is more complex compared to single-line text recognition. To address multi-line text recognition, a text detection model can first detect text lines within the paragraph, such as \cite{tian2016detecting, bera2021distance, kundu2020text}, and then a text line recognition model can be used to recognize each text line, such as \cite{yu2024approach}, which combines implicit and explicit segmentation to recognize text lines; \cite{kang2022pay}, which uses a transformer model for text line recognition; \cite{wang2020writer}, which proposes a Multi-Branch guided Attention Network (MBAN) to improve the accuracy of recognizing irregular text sequences; and \cite{diao2024hierarchical}, which proposes a novel interaction manner namely hierarchical visual-semantic interaction (HVSI) for text line recognition. There are also studies that combine these two aspects into a unified approach. For example, \cite{wigington2018start,tensmeyer2019training} use a Region Proposal Network to find the Start-Of-Line (SOL), followed by an RNN using SOL to segment normalized line text images. Finally, an advanced CNN-LSTM model is used to predict the text encoding sequence. \cite{moysset2017full} predicts SOL and End-Of-Line (EOL) to locate the regions of the text lines. \cite{chung2019computationally} also divides the entire recognition task into two stages: the first stage is text localization, and the second stage is text recognition.

As research in this field has progressed, other methods for accomplishing paragraph text recognition have emerged. For example, \cite{huang2019adversarial} uses adversarial models to enhance recognition accuracy, while \cite{peng2022pagenet} employs finer segmentation granularity, treating each character as a target for localization and recognition. Alternatively, some approaches forgo explicit segmentation and instead achieve implicit segmentation during the training process of the recognition model, resulting in what are known as implicit segmentation text recognition models, such as \cite{coquenet2021span}, which use identifiers to distinguish between different text lines. Going a step further, models like \cite{yousef2020origaminet} can differentiate between text lines using only text transcriptions. One benefit of this approach is that it reduces the need for segmentation annotations, which require significant manual works.

Language differences impact the approach to HPTR, especially when comparing Chinese to Western languages. In Western languages, words and text lines are composed of characters arranged sequentially, allowing paragraph text recognition to be treated as a sequence recognition problem. This means that recognizing each character with high precision is less critical. As a result, models like \cite{yousef2020origaminet, kumari2023comprehensive, singh2021full} employ line-level or multi-line segmentation to recognize paragraph text images. 
In contrast, Chinese presents greater challenges due to the large number of characters and the rich two-dimensional information inherent in the language. These features make recognition more difficult, requiring more powerful recognition capabilities to classify each character accurately. Models such as \cite{ma2020joint, peng2022pagenet} rely on character-level segmentation annotations to train the model effectively. Additionally, there has been research into multilingual recognition models, such as \cite{chen2020multrenets,lundgren2019octshufflemlt}, which aim to improve the generalizability of the model across different languages, thereby enhancing its practical value.

\subsection{Structural Re-parameterization}
\label{ssc:sr}
The key of Structural Re-parameterization is to use the parameters of one structure to parameterize those of another structure. \cite{ding2021resrep} utilizes re-parameterization to design a pruning method called ResRep. Compared to previous pruning methods, ResRep not only achieves a high compression ratio but also maintains high accuracy in the pruned model. \cite{ding2021repvgg} proposes a network model that utilizes structural re-parameterization to decouple of the inference-stage and the training-stage architecture, which has a VGG-like inference-stage body consisting solely of a stack of $3 \times 3$ convolutions and ReLU layers, while the training-stage model has a multi-branch topology. \cite{ding2021diverse} propose a universal building block for Convolution Neural Networks to enhance performance without any inference-time costs. This block, called the Diverse Branch Block (DBB), boosts the feature representational capacity of a single convolution by incorporating diverse branches of different scales and complexities to enrich the feature space. These branches include sequences of convolutions, multi-scale convolutions, and average pooling. A DBB can be equivalently converted into a single-convolution layer for deployment after training. \cite{ding2021repvgg} introduce a structural re-parameterization technique that enhances the power of fully connection (FC) layers for image recognition by incorporating local priors. \cite{ding2019acnet} discovers that parameters located at the skeleton of convolution kernels are more crucial than those at other positions. Therefore, through structural re-parameterization, during the training stage, they training the parameters of the $K \times 1$ convolution kernel and the $1 \times K$ convolution kernel to reinforce the parameters of the K × K convolution kernel skeleton.

\section{Architecture}
\label{sc:architecture}

This section will introduce the structural design of RVAFM concerning structural re-parameterization. First, Section \ref{ssc:backbone} introduces the VAN model, providing a clearer illustration of the procedure of RVAFM. Next, Section \ref{ssc:repvam-training} explains how to modify VAM using dual-parameter layers to obtain RVAFM during the training stage. Thirdly, Section \ref{ssc:conversion} demonstrates how RF seamlessly transform RVAFM from its training stage structure to its inference stage structure using the additivity distributive properties of convolution operations and matrix multiplication. Finally, Section \ref{ssc:analysis} analyzes from three perspectives why dual-parameter layers can enhance the learning capability of RVAFM. 

For understanding the following content, Tab. \ref{tb:layer-symbol} lists the symbols representing the parameter layers in RVAFM, which corresponds of Fig. \ref{fig:arch-repvamuyb}.

\begin{table}[!t] \normalsize
\centering
\renewcommand\arraystretch{1.25}
	\caption{the symbols of the parameter layers in RVAFM}
	\label{tb:layer-symbol}
  \setlength{\tabcolsep}{3.5mm}{
	\begin{tabular}{cc}
		\hline
		 \textbf{Symbols} & \textbf{Represent} \\
   \hline
 $D_{h}^{d}$& The dual-dense layer consisting of $D_{h}^{(1)}$ and $D_{h}^{(2)}$ \\
 $D_{f}^{d}$& The dual-dense layer consisting of $D_{f}^{(1)}$ and $D_{f}^{(2)}$ \\
 $D_{j}^{d}$& The dual-dense layer consisting of $D_{j}^{(1)}$ and $D_{j}^{(2)}$ \\
		 $D_{a}^{d}$& The dual-dense layer consisting of $D_{a}^{(1)}$ and $D_{a}^{(2)}$ \\
 $F^{d}$& The dual-convolution layer consisting of $F^{(1)}$ and $F^{(2)}$\\
 \hline
 $D_{h}$&The single-dense layer corresponding to $D_{h}^{d}$\\
 $D_{f}$&The single-dense layer corresponding to $D_{f}^{d}$\\
 $D_{j}$&The single-dense layer corresponding to $D_{j}^{d}$\\
 $D_{a}$&The single-dense layer corresponding to $D_{a}^{d}$\\
 $F$&The single-convolution layer corresponding to $F^{d}$\\
 \hline
	\end{tabular}
 }
\end{table}

\subsection{Backbone}
\label{ssc:backbone}

Before we delve into RVAFM, it is important to provide a brief overview of the VAN model to better understand the procedure of RVAFM.
The architecture of VAN consists of an encoder, a VAM, and a decoder. This explanation will clarify the input and output of the RVAFM, and its role within the entire model. The following describes the procedure of VAN:

The encoder of VAN comprises a substantial composition of convolution Blocks (CB) and Depthwise Separable convolution Blocks (DSCB). It takes an paragraph text image \(X \in \mathbb{R} ^{H \times W \times C} \) as input, where \(H\), \(W\), and \(C\) represent height, width, and the number of channels respectively. Then, it outputs the feature map \(f \in \mathbb{R} ^{H_{f} \times W_{f} \times C_{f}} \) with \(H_{f} = \frac{H}{32}\), \(W_{f} = \frac{W}{8}\) and \(C_{f} = 256\). 

VAM sequentially outputs the features of each line. At each stage, VAM includes four inputs: 1) The feature map \(f\) output by the encoder; 2) The weight \(\alpha_{t-1} \in \mathbb{R} ^{H_{f}} \) output from the previous stage; 3) The hidden state \(h_{t-1} \in \mathbb{R} ^{C_{h}} \) of the decoder from the previous stage; 4) The sum \(c_{t} \in \mathbb{R} ^{H_{f}} \) of all \(\alpha\) up to the current stage.
The feature map \(f\) firstly inputs AdaptiveMaxPooling, which compresses the width to a fixed value of 100. Then, a dense layer folds the horizontal dimension, resulting in the vertical representation \({f}' \in \mathbb{R} ^{H_{f} \times C_{f}}\).
The weight \(\alpha_{t-1}\) and the sum \(c_{t}\) (clamped between 0 and 1) are concatenated to form \(i_{t} \in \mathbb{R} ^{H_{f} \times 2} \), which is then passed through the one-dimensional convolution layer $F$ with \(C_{j} = 16\) filters of kernel size 15 and stride 1 to obtain \(j_{t} \in \mathbb{R} ^{H_{f} \times C_{j}}\). We define $\circledast$ as the convolution operation, $K$ represents the convolution kernel of $F$, and $b$ denotes the bias. Then, it can be expressed using the following formula: 
\begin{equation}
	\label{eq:jt}
	j_{t} = i_{t} \circledast K_{i} + b_{i}
\end{equation}

For each line \(i\), the previously obtained \({f_{i}}'\), \(j_{t, i}\) and \(h_{t-1}\) are processed through dense layers $D_{h}$, $D_{f}$, and $D_{j}$ respectively, to unify the channel number to \(C_{u} = 256\). Then, we sum them and apply the Tanh function to obtain multi-scale information \(s_{t,i} \in \mathbb{R} ^{C_{u}}\). This paper define the kernel of a dense layer as $W$. The formula can be expressed as follows: 
\begin{equation}
	\label{eq:st}
	s_{t} = tanh(f' \cdot W_{f}+ b_{f}+j_{t} \cdot W_{j}+ b_{j}+h_{t-1} \cdot W_{h}+ b_{h}
 )
\end{equation}
\begin{equation}
	\label{eq:tanh}
	tanh(x) = \frac{e^{x}-e^{-x}}{e^{x}+e^{-x}} 
\end{equation}

Eq. \ref{eq:st} contains both local and global information: the local information comes from the features and previous attention weights, while the global information comes from the decoder hidden state.
Input \(s_{t,i}\) into the dense layer $D_{a}$, compute a score \(e_{t,i}\) for each row feature. The formulas are shown as follows:
\begin{equation}
	\label{eq:et}
	e_{t} = s_{t} \cdot W_{a}+b_{a}
\end{equation}

The weight \(\alpha_{t}\) is obtain through softmax activation. Finally, $a_{t}$ is multiplied with the image feature map \(f\), and the result is summed along the vertical dimension to obtain the output \(l_{t}\). This represents the feature at time step \(t\), corresponding to the \(t\)-th row. 

Furthermore, the VAM applies \(s_{t}\) along with the current decoder hidden state \(h_{t}\) to obtain \(d_{t}\), which is used to predict whether the end of the paragraph has been reached.

The decoder of VAN takes \(l_{t}\) as input, and passes it through a single LSTM layer with \(C_{h} = 256\) to obtain another representation \(r_{t}\) of the same dimension. Subsequently, \(r_{t}\) undergoes a one-dimensional convolution layer that maps its channel dimension from 256 to \(N+1\) (where \(N\) represents the number of character types \(+\) one blank symbol). This process yields the final prediction \(p_{t}\) for that stage. CTC loss \citep{graves2006connectionist} is applied to align the $p_{t}$ with the corresponding line transcription $y_{t}$. All stage predictions \(p_{t}\) are combined to form the predicted result \(p_{pg}\) of the paragraph text image.

\subsection{RVAFM During The Training Stage }
\label{ssc:repvam-training}
From the previous section, one can notice that VAM uses a vertical attention mechanism to locate each text line by following the top-to-bottom order of the paragraph text image, marked by weight $a_{t}$, which identifies the text line at each stage and passes it to the decoder. To improve the accuracy of $a_{t}$ in locating text lines at each stage, this work modify VAM using two types of dual-parameter layers: dual-convolution layers and dual-dense layers, creating a new module called RVAFM, which decouples the structure of the training stage from the inference stage. In the training stage, RVAFM has a more complex, multi-branch, and multi-parameter structure to enhance the module's ability to learn features related to $a_{t}$, resulting in more precise implicit segmentation capabilities.

In this design, the dual-parameter layers, as the name implies, are modules composed of two parameter layers with identical configurations and equal status. Their procedures are similar: a basic structure receives the output from the previous module, which then becomes the input for the two complementary parameter layers. The outputs from these two parameter layers are summed, and the result is used as the output of this basic structure, which is then passed on to subsequent modules.

To obtain the structure of RVAFM in the training stage, we made modifications to three key computational points in VAM: The convolution layer $F$ is replaced by the dual-convolution layer $F^{d}$:
\begin{equation}
	\label{eq:djt}
	j_{t} = i_{t} \circledast K_{i}^{(1)} + b_{i}^{(1)} + i_{t} \circledast K_{i}^{(2)} +b_{i}^{(2)}
\end{equation}

The three dense layers $D_{h}$, $D_{f}$, and $D_{j}$ are replaced by three dual-dense layers $D_{h}^{d}$, $D_{f}^{d}$, and $D_{j}^{d}$:
 \begin{equation}
    \label{eq:dst}
    \begin{aligned}
     	s_{t} = tanh(&f' \cdot W_{f}^{(1)}+ b_{f}^{(1)}+f' \cdot W_{f}^{(2)}+b_{f}^{(2)}+  \\
                 &j_{t} \cdot W_{j}^{(1)}+ b_{j}^{(1)}+j_{t} \cdot W_{j}^{(2)}+b_{j}^{(2)}+  \\
                 &h_{t-1} \cdot W_{h}^{(1)}+ b_{h}^{(1)}+h_{t-1} \cdot W_{h}^{(2)}+b_{h}^{(2)}
 )
    \end{aligned}
\end{equation}

The dense layer $D_{a}$ is replaced by the dual-dense layer $D_{a}^{d}$:
\begin{equation}
	\label{eq:det}
	e_{t} = s_{t} \cdot W_{a}^{(1)} +b_{a}^{(1)}+ s_{t} \cdot W_{a}^{(2)}+b_{a}^{(2)}
\end{equation}

After making the above modifications, the new structure is used for the training stage of RVAFM. 

The final structure, as shown in Fig. \ref{fig:arch-repvamuyb}(a). The learning accuracy of RVAFM is improved through the multi-branch structure, but this also undoubtedly increases inference time. Therefore, a specific fusion method called Re-parameterization Fusion (RF) is proposed to fuse the model into a more lightweight structure, while ensuring that the accuracy remains consistent before and after the fusion. 

\begin{figure}[t]
    \centering
    \includegraphics[width=0.75\linewidth]{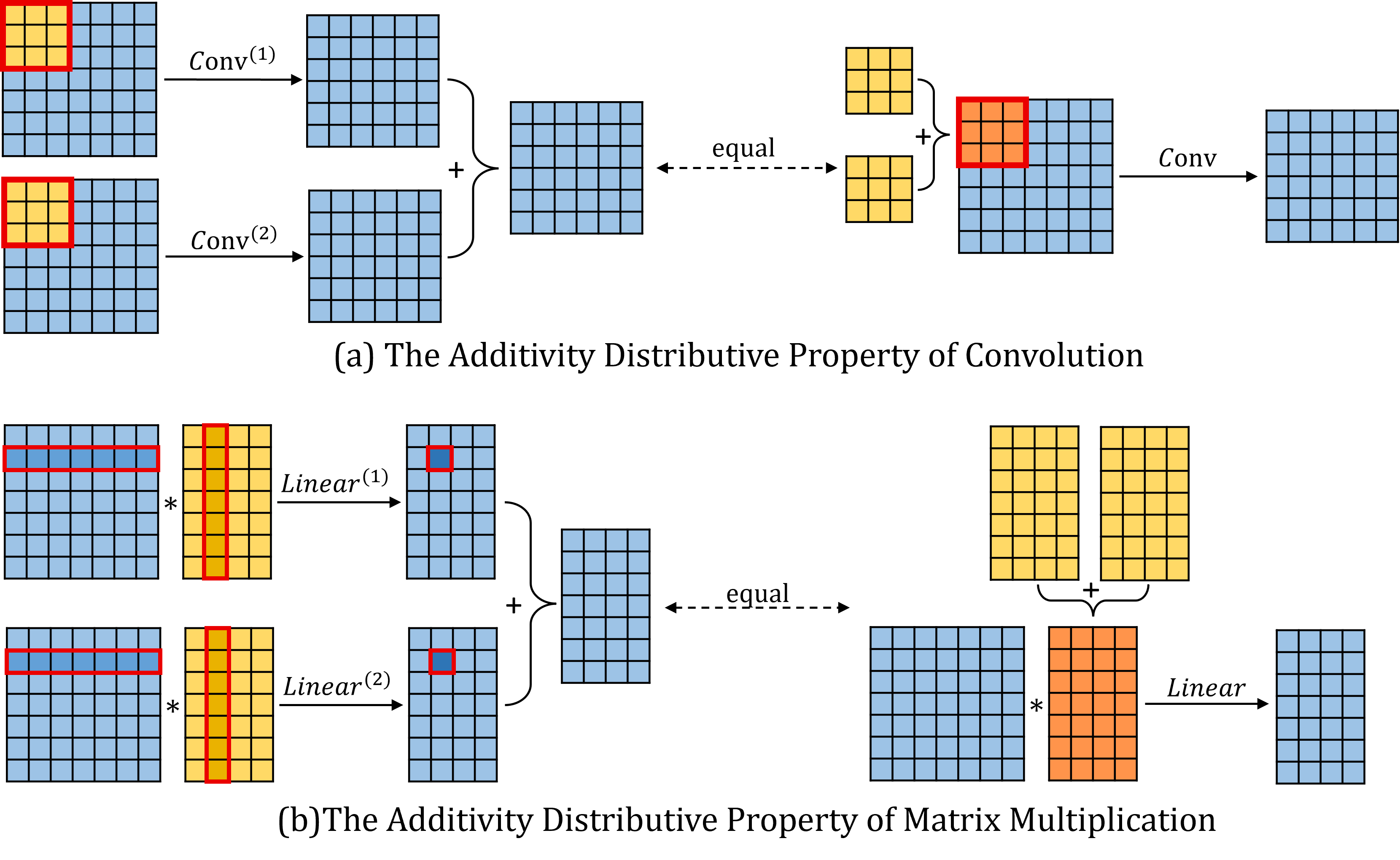}
    \caption{Visualization of the Additivity Distributive Property of Convolution and Matrix Multiplication}
    \label{fig:Visual-cm}
\end{figure}

\subsection{Fusion From Training-stage To Inference-stage}
\label{ssc:conversion}

Before using the trained RVAFM for the inference stage, RF is employed to lighten RVAFM. Specifically, RF includes two steps: Firstly, RF fuses the dual-parameter layers into single-parameter layers. Then, RF uses the parameters of the dual-parameter layers to parameterize the single-parameter layers. Importantly, it is ensured that this fusion process does not result in any loss of accuracy. The theorem and process of RF are explained in detail below. 

Firstly, the operations performed in the convolution layer can be represented by the following formula:
\begin{equation}
	\label{eq:conv}
	O = I \circledast K + b
\end{equation}
where $I$ represents the input feature map of the convolution layer, and $O$ represents the output of the convolution layer. Specifically, the convolution operation involves element-wise multiplication of a convolution kernel with a sliding window moving over the input feature map according to a fixed stride, followed by summation to obtain the result. In this process, involving only multiplication and addition operations, it can be easily shown the linearity of convolution operations, which exhibit the additivity distributive property, expressed as follows: 
 \begin{equation}
 \label{eq:conv-add}
 \begin{aligned}
    I  \circledast& K^{(1)} + b^{(1)} + I \circledast K^{(2)} + b^{(2)} \\
    &=  I \circledast (K^{(1)} + K^{(2)}) + (b^{(1)} + b^{(2)})
    \end{aligned}
 \end{equation}

The visualization of the additivity distributive property of Convolution is shown in Fig. \ref{fig:Visual-cm}(a). According to this theorem, we first create a single-convolution layer with the same configuration as the dual-convolution layers. Then, we parameterize the single-convolution layer by summing the parameters of the dual-convolution layers. Finally, the single-convolution layer is used for inference. This accomplishes the fusion of the dual-convolution layer $C^{d}$, corresponding to Eq. \ref{eq:djt}. 

Next, for the dense layer, its operations can be represented by the following formula: 
\begin{equation}
	\label{eq:linear}
	O = I \cdot W + b
\end{equation}

Specifically, the matrix multiplication performed by the dense layer involves only multiplication and addition operations. Therefore, it is easy to prove its linearity, which includes the additivity distributive property, as shown below: 
\begin{equation}
	\label{eq:linear-add}
    \begin{aligned}
        I \cdot &W^{(1)} + b^{(1)} + I \cdot W^{(2)} + b^{(2)} \\
        &= I \cdot(W^{(1)} + W^{(2)}) +(b^{(1)} + b^{(2)})
    \end{aligned}
\end{equation}

The visualization of the additivity distributive property of Matrix Multiplication is shown in Fig. \ref{fig:Visual-cm}(b). According to this theorem, we can similarly parameterize a new single- layer by summing the parameters of the dual-dense layers, just as with the dual-convolution layer. This approach allows us to fuse the four dual-dense layers $D_{h}^{d}$, $D_{f}^{d}$, $D_{j}^{d}$ and $D_{a}^{d}$ , corresponding to Eq. \ref{eq:dst} and Eq. \ref{eq:det}. 

Finally, the inference stage RVAFM, as shown in Fig. \ref{sc:architecture}(b), features a single branch, requiring less storage and computation. Thereby, it is simpler and lighter compared to before the fusion.
RF undoubtedly reduces the computational load of the module and ensures inference efficiency. Significantly, according to Eq. \ref{eq:conv-add} and Eq. \ref{eq:linear-add}, the overall operation of RVAFM before and after the fusion are equivalent, which means there is no loss in the module accuracy.

\subsection{The Analysis For The Dual-parameter Layer}
\label{ssc:analysis}
For the design of RVAFM, the dual-parameter layer serves as a practical implementation of structural re-parameterization technology with two primary objectives: 1) Adapting RVAFM into a multi-branch structure during the training stage. 2) Ensuring that the inference efficiency of RVAFM remains irrelevant to the multi-branch structure through RF. This subsection examines how the dual-parameter layer enhances the learning effectiveness of RVAFM from three distinct perspectives.

Firstly, during the training stage, every deep learning system has a chance to fail training for various reasons, which can be classified into two types: traceable reasons and untraceable reasons. Specifically, the former can be addressed through specialized methods, such as significant noise in the training samples, which can be solved by using denoising techniques. However, the latter, due to their untraceable nature, cannot be resolved through specific methods and can only be mitigated by employing general strategies to ensure the training accuracy. A simple general strategy is to train multiple systems simultaneously and then determine the prediction based on the combined outputs of all systems, The principle of the dual-parameter layer we use is consistent with this strategy.
From the perspective of the model system, the single-parameter layer can be viewed as a small system, while the dual-parameter layer can be seen as a central system that combines two identical smaller systems through an addition operation, sharing the output from the previous module. Both systems may fall short of achieving the desired training objectives. Nevertheless, the central system demonstrates a better robustness because if one small system encounters an issue, the other can compensate. Moreover, the likelihood of both systems failing simultaneously is lower than that of a single system. 

Secondly, in the training process of a deep learning model, the information learned is propagated back through the network via backpropagation. This process updates the parameters by gradients, aiming to minimize the difference between the predicted outputs and the ground truth labels. The single-branch structure can result in a situation where errors in the learning of the later parameter layers can affect the learning of the earlier parameter layers. 
In contrast, the models with a multi-branch structure can effectively mitigate this issue. More specifically, when the parameter layers of one branch encounter learning errors, the corresponding parameter layers of other branches can compensate for these deficiencies, thereby reducing the impact on earlier layers.
From the perspective of information transmission, a single-parameter layer has only one pathway for propagating the learned information, whereas a dual-parameter layer features two branches with equal status. RVAFM, which integrates five dual-parameter layers, consequently has $2^{5}$ branches. These branches are not entirely independent, as a node issue may impact multiple branches passing through it. However, even in such cases, the remaining branches can still convey accurate information. In contrast, a single-branch structure would experience complete disruption in such scenarios. As discussed, a multi-branch structure increases the number of pathways for information flow during backpropagation, thereby significantly enhancing the stability of the model's learning process.

Thirdly, by analyzing from the functionality of a deep learning model, it becomes clear that the overall functionality is composed of the functions of all parameter layers. We can consider the learning process of the model as each parameter layer developing its specific function. Each parameter layer in this training process has two training directions: 1) to broaden the function, 2) to deepen the function. In the early stages of training, all parameter layers can be considered as having no function. At this point, the learning direction focuses on the former, allowing each parameter layer to establish its own function from scratch. In the the middle and later stages of training, the learning direction gradually shifts to the latter, strengthening each parameter layer's function to enhance the overall functionality of the model.
From the perspective of training the parameter layer's function, Compared to directly training a single-parameter layer, we first train a dual-parameter layer, which enhances the learning effectiveness of the parameter layer from both directions mentioned above. In training, the dual-parameter layer encounters two scenarios: 1) when the two sub-layers within the dual-parameter layer are establishing different functions, which enhances the layer's function in term of breadth, and 2) when both layers are evolving towards the same function, which improves the layer's function in term of depth. Ultimately, when the dual-parameter layer is fused into a single-parameter layer, the single-parameter layer benefits from improved functionality, leading to increased model accuracy. 

\section{Experiments}
\label{sc:experiments}
This section is divided into four parts: Section \ref{ssc:ec} introduce the experimental conditions. Section \ref{ssc:csoa} demonstrates that the recognition capability of RVAN has reached an advanced level by comparing it with a series of state-of-the-art HPTR models. Section \ref{ssc:cvan} further proves, through comparison with VAN, that the structural re-parameterization technique not only provides superior learning ability but also maintains the inference efficiency. Section \ref{ssc:pte} studies the hyper-parameters for RVAFM to find the optimal setting. Section \ref{ssc:ae} presents ablation experiments on all dual-parameter layers, aimed at exploring their contributions and interconnections. 

\subsection{Experimental Conditions}
\label{ssc:ec}
\begin{figure}[t]
    \centering
    \includegraphics[width=0.5\linewidth]{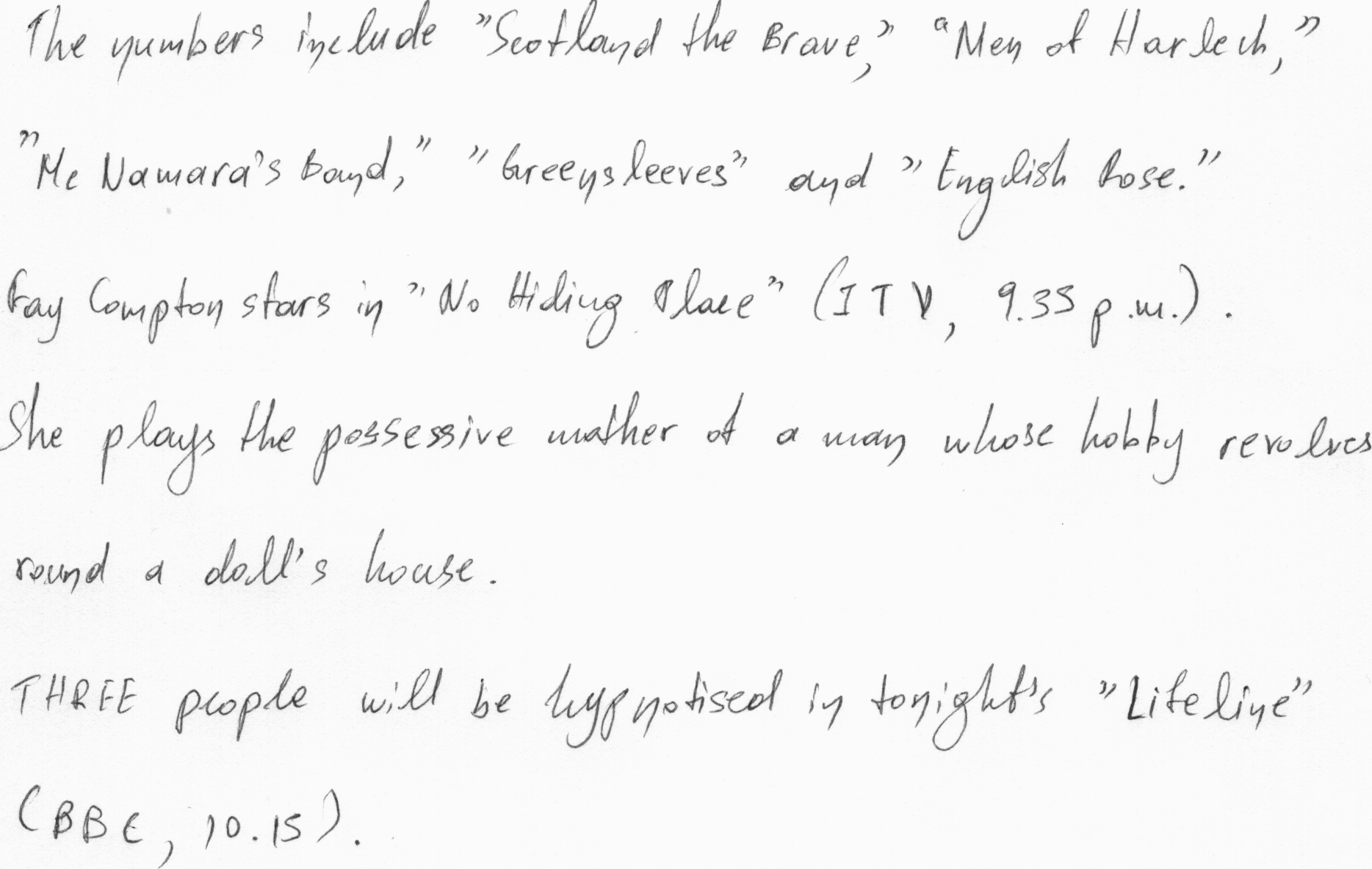}
    \caption{A sample image from the IAM paragraph-level dataset }
    \label{fig:Visual-iam}
\end{figure}
In this paper, we use the IAM dataset\citep{marti2002iam}, which consists of handwritten copies of text paragraphs extracted from the LOB corpus. It comprises gray-scale images of English handwriting at a resolution of 300 dpi. The dataset provides segmentation and corresponding transcriptions at the page, paragraph, line, and word levels. In this experiment, we use the paragraph level with a common but unofficial split, resulting in the following number of images in each dataset: 747 in the training set, 116 in the validation set, and 336 in the test set. Fig. \ref{fig:Visual-iam} shows a sample image from the IAM paragraph-level dataset. 

For pre-processing, this work employs bilinear interpolation to downscale the input images by a factor of 2, resulting in images from the IAM dataset with a resolution of 150 dpi. Additionally, zero-padding is applied to ensure that the input images meet the minimum height of 480 pixels and minimum width of 800 pixels. This ensures a minimum feature width of 100 and a minimum feature height of 15, as required by RVAN and VAN. 

For data augmentation, we use the following techniques during training to reduce over-fitting: elastic distortion, perspective transformation, resolution modification, random projective transformation, erosion and dilation, brightness and contrast adjustment, and sign flipping. Each transformation is applied with a probability of $20\%$. It is important to note that elastic distortion, perspective transformation, and random projective transformation are mutually exclusive, and the other techniques can be applied sequentially in the given order. 

For metrics, we use Character Error Rate (CER) and Word Error Rate (WER) as the standards for assessing model accuracy. These metrics are calculated based on the Levenshtein distance between the ground-truth text $y$ and the recognized text $\hat{y}$, normalized by the length of the ground truth $len(y)$. The formula for CER is as follows: 
\begin{equation}
	\label{eq:cer}
	CER=\frac{\sum_{i=1}^{K} Levenshtein(\hat{y_{i}},y_{i})}{\sum_{i=1}^{K}len(y_{i})}
\end{equation}

In Eq. \ref{eq:cer}, $K$ represents the number of paragraph text images in the dataset. The formula for WER is similar to that of CER but is calculated at the word level. 

For the training configuration, we use the PyTorch framework and incorporate the apex package to enable mixed-precision training, which reduces memory consumption and improves training speed. We use the Adam optimizer with an initial learning rate of $10^{-4}$. Training and inference are performed on a single GeForce RTX 3090 GPU with a mini-batch size of 8. 

\subsection{Comparison with State-of-the-Art Handwritten Paragraph Text Recognition Method}
\label{ssc:csoa}

Tab. \ref{tb:comparative-soa} presents the results of current state-of-the-art HPTR models compared to RVAN on the IAM test set, focusing on CER and WER. \citep{coquenet2022end} represents the original VAN, which serves as the baseline model for RVAN. The comparison reveals that our RVAN outperforms VAN in both CER and WER metrics, which demonstrates that RVAFM effectively enhances the accuracy of VAN. 

Additionally, the best results in Tab. \ref{tb:comparative-soa} are highlighted in bold. It can be observed that \cite{bluche2015deep} achieves the state-of-the-art results in both CER and WER metrics, 
which is primarily due to leveraging the advantages of multiple systems. However, our proposed method achieves the same level in the CER metric and surpasses all other methods except \cite{bluche2015deep} in the WER metric, which demonstrates that our proposed method is promising. 

\begin{table}[!t] \normalsize
\centering
\renewcommand\arraystretch{1.25}
	\caption{Comparison with State-of-the-Art Paragraph-Level Recognition Methods on the IAM test set }
	\label{tb:comparative-soa}
  \setlength{\tabcolsep}{3.5mm}{
	\begin{tabular}{l|cc}
		\hline
		 \textbf{Method}& \textbf{CER}(\%)& \textbf{WER}(\%)\\
   \hline
 Carbone et al. \cite{carbonell2019end}& $15.6$&/\\
 Chung et al. \cite{chung2019computationally}& $8.5$&/\\
 Huang et al. \cite{huang2019adversarial}& $7.1$&$21.9$\\
		 Wigington et al. \cite{wigington2018start}& $6.4$& $23.2$\\
 Bluche \cite{bluche2016joint}& $7.9$&$24.6$\\
 Bluche et al. \cite{bluche2017scan}& $16.2$&/\\
 Bluche \cite{bluche2015deep}& $\textbf{4.4}$&$\textbf{10.9}$\\
 Yousef et al. \cite{yousef2020origaminet}& $4.7$&/\\
 Coquenet et al. \cite{coquenet2021span}& $5.45$&$19.83$\\
 Coquenet et al. \cite{coquenet2022end}& $4.45$&$14.55$\\
 \hline
 Our& $\textbf{4.4}$&$14.37$\\
 \hline
	\end{tabular}
 }
\end{table}

\subsection{Comparison with the original VAN }
\label{ssc:cvan}

This subsection will provide a more detailed comparison between RVAN and VAN. In this part, we employed three models: RVAN-T, which represents RVAN before fusion; RVAN-I, which represents RVAN after fusion; and VAN. Apart from the differences in the VAM, the encoder and decoder configurations are identical across all three models to ensure the accuracy and validity of the comparative experiment. These three models are trained and tested on the IAM training set and test set, respectively. We records their CER, WER, model parameter count, and the processing time per sample (with the model set to inference mode), as shown in Tab. \ref{tb:comparative}. 

The results of RVAN-T and RVAN-I show that the accuracy of RVAN remains consistent before and after the fusion. 
Additionally, the number of parameters of the model and the time required to process a single sample both decrease after the fusion. This consistency between practical results and the theoretical framework presented in Section \ref{ssc:conversion} confirms that our designed RVAFM can achieve lossless fusion. This allows RVAFM to retain all the information learned during training while simplifying the model structure. 

The results of RVAN-T and VAN show that modifying VAN with dual-parameter layers into a multi-branch structure enhances its accuracy, but this comes at the expense of inference speed. In contrast, the results of RVAN-I and VAN show that RVAN-I not only retains the high accuracy of RVAN-T but also eliminates the negative impact on inference speed through RF. 
In summary, the structural re-parameterization technique proves to be highly effective when applied to VAM. It decouples the model's structure during the training and inference stage, allowing each to excel in its specific role. The training stage employs a complex multi-branch structure to learn more effective information, while the inference stage uses a simpler single-branch structure to accelerate the inference speed. Importantly, the conversion between the two structures results in no loss of accuracy. 
\begin{table}[!t]\normalsize
\centering
\renewcommand\arraystretch{1.25}
	\caption{Comparison with the original VAN on the IAM test set}
	\label{tb:comparative}
  \setlength{\tabcolsep}{1.7mm}{
	\begin{tabular}{l|cccc}
		\hline
		\textbf{Model}& \makecell[c]{\textbf{CER}\\ (\%)}& \makecell[c]{\textbf{WER} \\ (\%)}& \makecell[c]{\textbf{Parameters} \\ (MB)}& \makecell[c]{\textbf{Sample Time} \\ (ms)}\\
        \hline
		VAN& $4.45$& $14.55$& $2.7$ & $19.3$\\
		RVAN-T& $\textbf{4.40}$& $\textbf{14.37}$& $2.7$& $19.6
$\\
		RVAN-I& $\textbf{4.40}$& $\textbf{14.37}$& $\textbf{2.6}$& $\textbf{18.3}
$\\
		\hline
	\end{tabular}
 }
\end{table}
\begin{table}[!t]\normalsize
\centering
\renewcommand\arraystretch{1.5}
	\caption{The results of RVAN models with different $C_{u}$ values}
	\label{tb:PT-Cu}
  \setlength{\tabcolsep}{3.9mm}{
	\begin{tabular}{l|ccc}
		\hline
		\textbf{Model} &\makecell[c]{\textbf{CER} \\(\%)}& \makecell[c]{\textbf{WER} \\ (\%)}& \makecell[c]{\textbf{Sample Time} \\ (ms)} \\
        \hline
		RVAN-Cu64&$4.53$& $14.82$& $\textbf{18.0}$\\
		RVAN-Cu128&$4.55$& $15.08$& $18.1$\\
		RVAN-Cu256&$\textbf{4.46}$& $\textbf{14.69}$& $18.3
$\\
 RVAN-Cu512&$5.01$& $16.21$&$19.8
$\\
        \hline
	\end{tabular}
 }
\end{table}

\subsection{Parameter Tuning Experiments }
\label{ssc:pte}

\begin{figure}[t]
    \centering
    \includegraphics[width=0.75\linewidth]{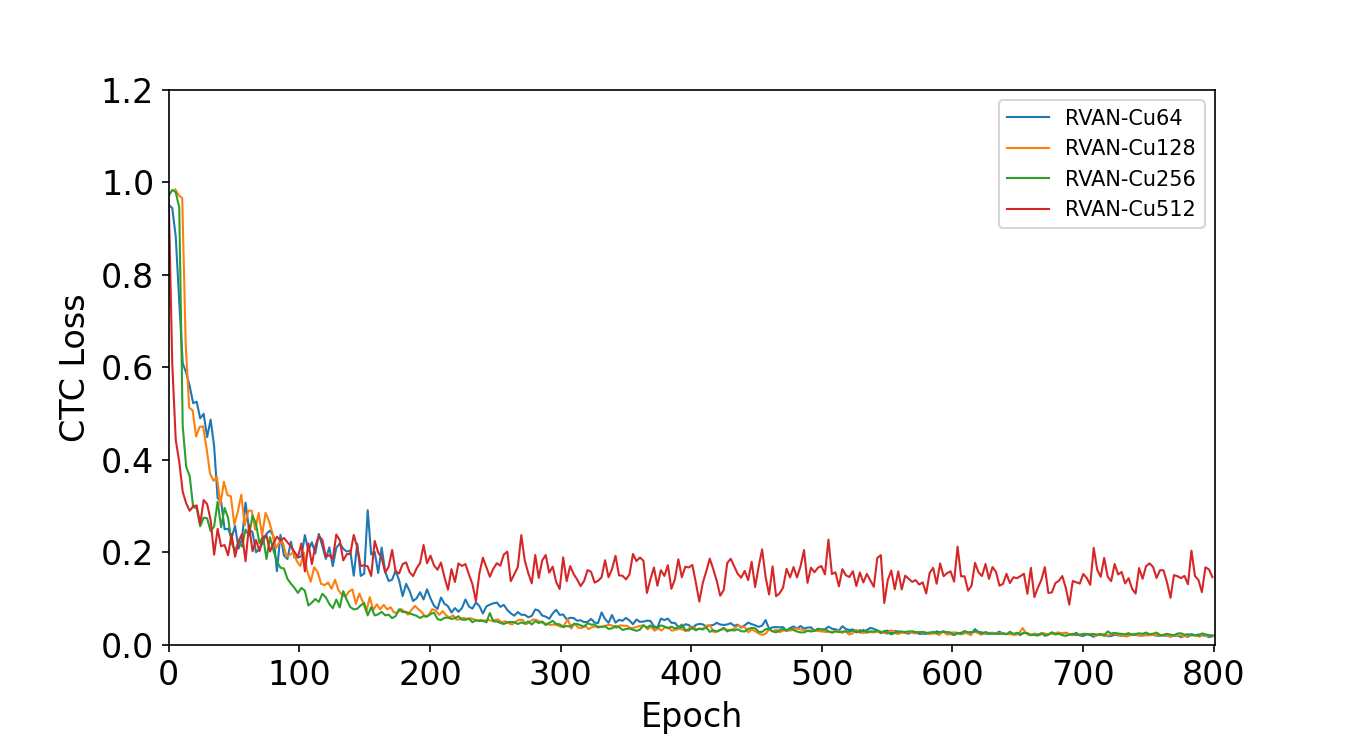}
    \caption{The training curves of RVAN models with different $C_{u}$ values }
    \label{fig:Visual-Cu-ctcloss}
\end{figure}

This subsection conducts a series of hyper-parameter tuning experiments to find the most suitable parameters for RVAFM. $C_{u}$ and \textit{the number of sub-layers} (referred to as NSL) are the hyper-parameters studied in this work. Specifically, $C_{u}$ represents the channel number for attention. With increasing $C_{u}$, the dimensions of the four dual-dense layers and the size of the intermediate feature $s_{t}$ will also increase, whereas decreasing $C_{u}$ will have the opposite effect, both of which directly impact the learning ability and inference speed of RVAFM. NSL refers to the number of identical sub-layers within a multi-parameter layer. The proposed RVAFM is primarily constructed using multi(dual)-parameter layers, specifically with an NSL of 2. 
Modifying NSL directly affects the pathways that propagate the learned information and the parameters that need to be learned. 

For the hyper-parameter tuning experiment of $C_{u}$, we employed four models: RVAN-Cu64, RVAN-Cu128, RVAN-Cu256, and RVAN-Cu526, where the numbers at the end represents the value of $C_{u}$. 
During training, the four models differ only in their $C_{u}$ values, and all other training configurations are kept identical. 
We train the four models for 800 epochs and record their training curves and performance on the test set. Based on these results, we select the appropriate $C_{u}$. 

The training curves for the experiment involving $C_{u}$ are shown in Fig. \ref{fig:Visual-Cu-ctcloss}. As can be seen from the figure, although RVAN-Cu512 has the large $C_{u}$ value, it does not necessarily result in the best training performance among the four models. This may be due to the fact that a larger $C_{u}$ value brings stronger expressive power but also makes the learning task more challenging. For the other three models, their loss has been stabilized after $500$ epochs. Among them, the training curve of the RVAN-Cu256 model is lower than that of the other models, indicating that with $C_{u}=256$, the model achieves relatively better convergence among the four values. 
As illustrated, the training curve of RVAN-Cu512 exhibits the fastest decline before approximately 50 epochs. However, after 50 epochs, it fluctuates around a high loss value, while the curves of the other three models continue to decrease. Among them, RVAN-Cu256 demonstrates the best performance with a fast and stable convergence. According to the testing results in Tab. \ref{tb:PT-Cu}, in terms of accuracy, RVAN-Cu256 also performs the best. In terms of inference speed, although RVAN-Cu64 is faster, the difference compared to RVAN-Cu256 is not obvious. Balancing accuracy and efficiency, we select $256$ as the optimal $C_{u}$ for RVAN. 

\begin{figure}[t]
    \centering
    \includegraphics[width=0.75\linewidth]{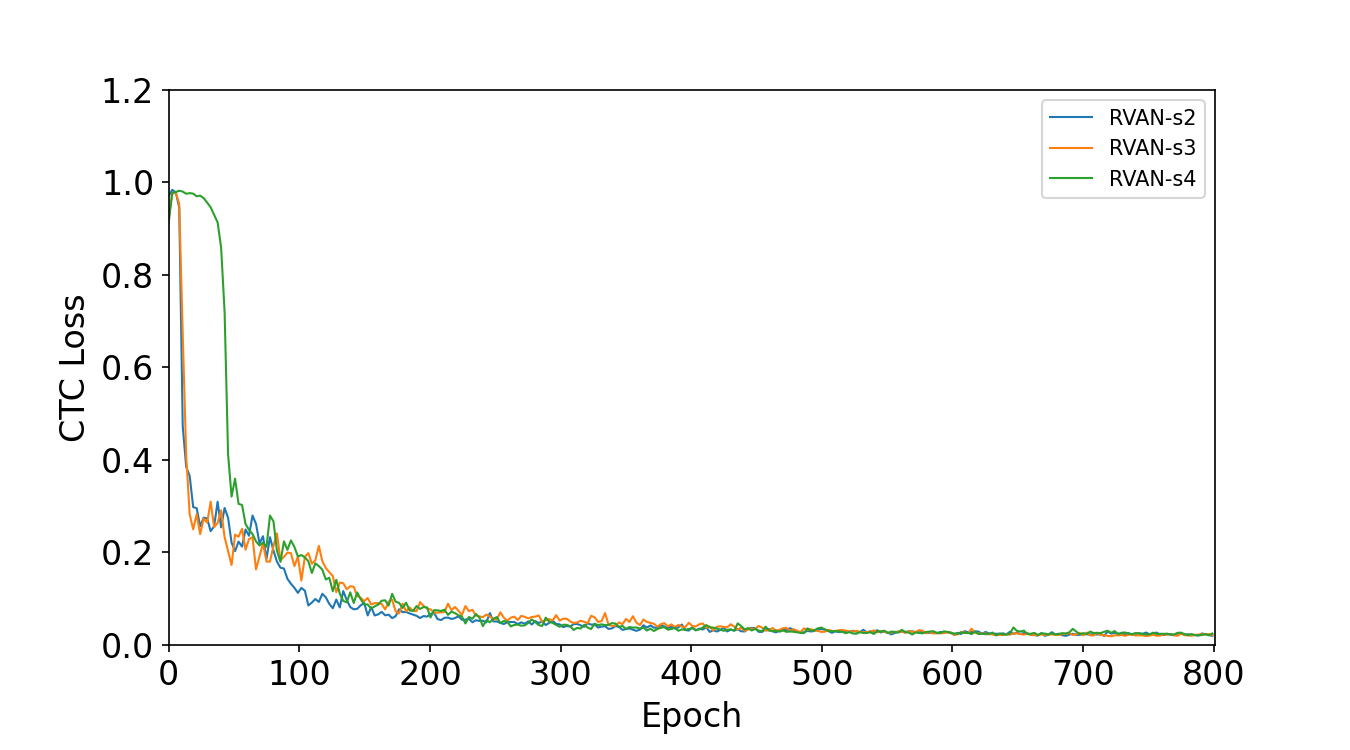}
    \caption{The training curves of RVAN models with different NSL}
    \label{fig:Visual-s-ctcloss}
\end{figure}

\begin{table}[!t]\normalsize
\centering
\renewcommand\arraystretch{1.5}
	\caption{The results of RVAN models with different numbers of sub-parameter layers}
	\label{tb:PT-s}
  \setlength{\tabcolsep}{3.9mm}{
	\begin{tabular}{l|cc}
		\hline
		\textbf{Model} &\makecell[c]{\textbf{CER} \\(\%)}& \makecell[c]{\textbf{WER} \\ (\%)}\\
        \hline
		RVAN-s2&$\textbf{4.46}$& $\textbf{14.69}$\\
		RVAN-s3&$4.54$& $14.89$\\
 RVAN-s4&$4.67$& $15$\\
        \hline
	\end{tabular}
 }
 \end{table}
For the hyper-parameter tuning experiment of NSL, we apply three models: RVAN-s2, RVAN-s3, and RVAN-s4, where the numbers represent the number of sub-layers in the five multi-parameter layers. Specifically, RVAN-s2 represents the proposed RVAN, utilizing dual-parameter layers. RVAN-s3 and RVAN-s4 are built on the same pattern. It is important to note that the three models are identical in configuration, except for NSL. We train the three models for 800 epochs and record their training curves as well as the best CER and WER on the test set. Finally, based on this evidence, we select the most suitable hyper-parameter.

Fig. \ref{fig:Visual-s-ctcloss} shows the training curves of the RVAN models with three different NSL. From this figure, it can be observed that the performance of the three models is quite similar. However, the training curve of RVAN-s2 is slightly better than the other two curves, indicating that increasing NSL further results in limited or even no improvement in model performance. This conclusion is also supported by Tab. \ref{tb:PT-s}, where the final performance of RVAN-s3 and RVAN-s4 on the test set is inferior to that of RVAN-s2. We attribute this to two main reasons: 1) larger models require more training epochs to fully realize their potential, and 2) simply increasing NSL cannot infinitely improve model accuracy, as other factors may also constrain the model's performance. Therefore, to balance efficiency and accuracy, selecting NSL as 2 is the optimal choice of hyper-parameter. 

\subsection{Ablation Experiments}
\label{ssc:ae}
In this subsection, we conduct controlled experiments on the five dual-parameter layers. VAN model is used as the baseline. Based on the five dual-parameter layers, we develop six variant models: five models, each with a different dual-parameter layer to assess their individual effectiveness, and one model incorporating all dual-dense layers to explore the model's performance without dual-convolution layers. For comparability and fairness, all models are trained under the same configuration for 800 epochs, and the results are recorded in Tab. \ref{tb:ablation}

By observing Tab. \ref{tb:ablation}, we can see that all variants have improved the accuracy of the baseline model to some extent, although they still fall short compared to RVAN under the same configuration. There are two particularly interesting points to note:
\begin{enumerate}
    \item Among the variants that use only one dual-parameter layer, the one utilizing $D_{j}^{d}$ demonstrates the best accuracy suggesting that there is substantial potential for further exploration in the feature information that includes the vertical attention weights from all previous stages. We only use this dual-parameter layer as a more efficient choice, particularly when training resources are limited. 
    \item The variant that incorporates all dual-dense layers shows a noticeable improvement in accuracy compared to the variants where each dual-dense layer is applied individually. This indicates that there is a complementary relationship between the dual-dense layers. However, it still lags behind the RVAN that utilizes all dual-parameter layers, which underscores the significant impact that dual-convolution layers have on the performance of dual-dense layers.
\end{enumerate}

\begin{table}[!t]\normalsize
\centering
\renewcommand\arraystretch{1.5}
	\caption{The results of the ablation experiments}
	\label{tb:ablation}
  \setlength{\tabcolsep}{3.9mm}{
	\begin{tabular}{l|cc}
		\hline
		\textbf{Model} &\makecell[c]{\textbf{CER} \\(\%)}& \makecell[c]{\textbf{WER} \\ (\%)}\\
        \hline
		Baseline&$4.6$& $15.01$\\
		A) with $(W_{h}^{(1)},W_{h}^{(1)})$&$4.55$& $14.81$\\
 B) with $(W_{f}^{(1)},W_{f}^{(1)})$&$4.56$& $14.9$\\
 C) with $(W_{j}^{(1)},W_{j}^{(1)})$& $4.52$&$14.74$\\
 D) with $(W_{a}^{(1)},W_{a}^{(1)})$& $4.56$&$14.8$\\
 E) with $(F_{i}^{(1)}, F_{i}^{(2)})$& $4.57$&$14.84$\\
 F) with all dual-dense layers& $4.51$&$14.7$\\
 RVAN& $4.46$&$14.69$\\
 \hline
	\end{tabular}
 }
 \end{table}

\section{Conclusion}
\label{sc:conclusion}

In this paper, based on VAM, we design RVAFM by incorporating structural re-parameterization techniques, which aims to decouple the structure of the module during the training and inference stage. During the training stage, RVAFM uses a multi-branch structure to achieve promising training performance. Before the inference stage, we employ the proposed fusion method RF to fuse the multi-branch structure into a single-branch structure while maintaining the learned capabilities, which ensures that the multi-branch structure does not affect the module inference speed. As a result, the model achieves further improvements in both recognition accuracy and inference efficiency. Our future research direction is to design a training strategy that dynamically adjusts the dual-parameter layers based on the states of different modules, thereby further improving the effectiveness of model training.

\section*{Acknowledgment}
This work was supported in part by the Guangdong Basic and Applied Basic Research Foundation under Grant 2022A515110020, the Jinan University ``Da Xiansheng'' Training Program under Grant YDXS2409, the National Natural Science Foundation of China under Grant 62271232, and the National Natural Science Foundation of China under Grant 62106085.
  \bibliographystyle{elsarticle-num-names} 
  \bibliography{main}



\end{document}